\documentclass[runningheads]{llncs}
\usepackage{graphicx}

\usepackage{pgfplots} % For plots
\usepackage{pgfplotstable} % For plots
\pgfplotsset{compat=newest} % For plots
\usetikzlibrary{plotmarks} % For plots
\usetikzlibrary{colorbrewer} % Nice colors

\usepackage{amsmath,amssymb}
\usepackage{tabularx}
\usepackage{multirow}
\usepackage{booktabs}
\usepackage{colortbl}
\usepackage{xcolor}
\DeclareMathOperator*{\argmax}{argmax}

\begin{document}
%===========================================================

\title{PReMVOS: Proposal-generation, Refinement and Merging for Video Object Segmentation} % Replace your paper's title here
\titlerunning{PReMVOS} % Replace an abstracted version of your paper's title here

%===========================================================

\author{Jonathon Luiten \and Paul Voigtlaender \and Bastian Leibe}
%
%Please include author names in full in the paper, 
%If any authors have names that can be parsed into FirstName LastName in multiple ways, please include the correct parsing, in a comment to the volume editors:
%\index{Lastnames, Firstnames}

\authorrunning{J. Luiten et al.} % A shorter version of authors' name

%===========================================================

\institute{
Computer Vision Group, RWTH Aachen University, Aachen, Germany\\
\email{\{luiten, voigtlaender, leibe\}@vision.rwth-aachen.de}
}

\maketitle

%===========================================================
\begin{abstract}
We address semi-supervised video object segmentation, the task of automatically generating accurate and consistent pixel masks for objects in a video sequence, given the first-frame ground truth annotations. Towards this goal, we  present the PReMVOS algorithm (Proposal-generation, Refinement and Merging for Video Object Segmentation). Our method separates this problem into two steps, first generating a set of accurate object segmentation mask proposals for each video frame and then selecting and merging these proposals into accurate and temporally consistent pixel-wise object tracks over a video sequence in a way which is designed to specifically tackle the difficult challenges involved with segmenting multiple objects across a video sequence. Our approach surpasses all previous state-of-the-art results on the DAVIS 2017 video object segmentation benchmark with a $\mathcal{J}$\&$\mathcal{F}$ mean score of 71.6 on the \texttt{test-dev} dataset, and achieves first place in both the DAVIS 2018 Video Object Segmentation Challenge and the YouTube-VOS 1st Large-scale Video Object Segmentation Challenge.

\end{abstract}
%===========================================================

\section{Introduction}

\begin{figure}[t!]
  \centering
    \includegraphics[width=10cm]{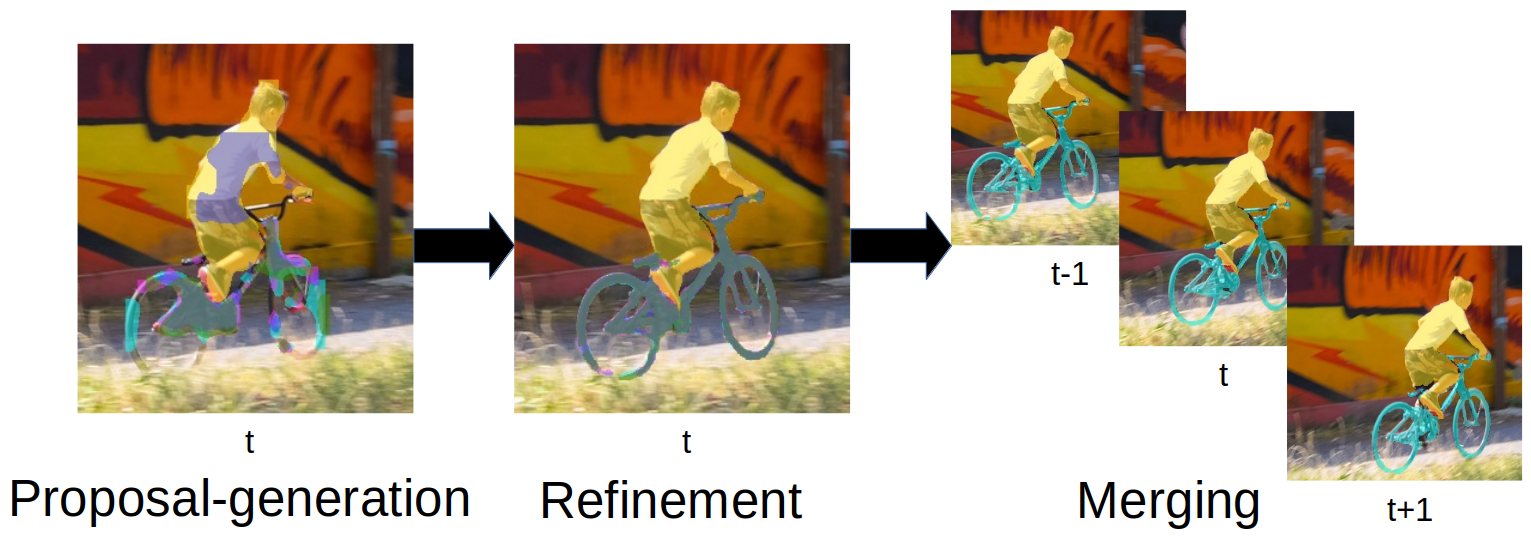}
  \caption{PReMVOS overview. Overlay colours represent different object proposals.}
  \label{fig:image}
\end{figure}

Video Object Segmentation (VOS) is the task of automatically estimating the object pixel masks in a video sequence and assigning consistent object IDs to these object masks over the video sequence. This can be seen as extension of instance segmentation from single frames to videos, and also as an extension of multi object tracking from tracking bounding boxes to tracking pixel masks. This framework motivates our work in separating the VOS problem into two sub-problems. The first being the instance segmentation task of generating accurate object segmentation mask proposals for all of the objects in each frame of the video, and the second being the multi object tracking task of selecting and merging these mask proposals to generate accurate and temporally consistent pixel-wise object tracks throughout a video sequence. Semi-supervised Video Object Segmentation focuses on the VOS task for certain objects for which the ground truth mask for the first video frame is given. The DAVIS datasets \cite{Caelles_arXiv_2018},\cite{Pont-Tuset_arXiv_2017},\cite{Perazzi2016} present a state-of-the-art testing ground for this task.
In this paper we present the PReMVOS (Proposal-generation, Refinement and Merging for Video Object Segmentation) algorithm for tackling the semi-supervised VOS task.  This method involves generating coarse object proposals using a Mask R-CNN like object detector, followed by a refinement network that produces accurate pixel masks for each proposal. We then select and link these proposals over time using a merging algorithm that takes into account an objectness score, the optical flow warping, a Re-ID feature embedding vector, and spatial constraints for each object proposal. We adapt our networks to the target video domain by fine-tuning on a large set of augmented images generated from the first-frame ground truth. An overview of our method, PReMVOS, can be seen in Figure \ref{fig:image}. Our method surpasses all current state-of-the-art results on all of the DAVIS benchmarks and achieves the best results in the 2018 DAVIS Video Object Segmentation Challenge \cite{DAVIS2018-Semi-Supervised-1st} and the YouTube-VOS 1st Large-scale Video Object Segmentation Challenge \cite{Luiten18ECCVW}.

%-------------------------------------------------------------------------

\section{Related Work}
Current state-of-the-art methods for VOS fall into one of two paradigms. The first is \textit{objectness} estimation with domain adaptation from first-frame finetuning. This approach, first proposed in \cite{caelles2017one}, uses fully convolutional networks to estimate the \textit{objectness} of each pixel by fine-tuning on the first-frame ground truth. 
This approach was expanded upon by \cite{maninis2017video} and \cite{voigtlaender17DAVIS},\cite{voigtlaender17BMVC} by using semantic segmentation guidance and iterative fine-tuning, respectively.
The second paradigm, used in several state-of-the-art methods \cite{Perazzi2017CVPR},\cite{li2017video},\cite{khoreva2017lucid},\cite{li2018video}, involves propagating the mask from the previous frame using optical flow and then refining these estimates using a fully convolutional network. The methods proposed in \cite{li2017video} and \cite{li2018video} expand this idea by using a network to calculate a re-identification (ReID) embedding vector for proposed masks and using this to improve the object re-identification after an object has been occluded. \cite{khoreva2017lucid} improves upon the mask propagation paradigm by training on a huge set of augmented images generated from the first-frame ground truth.
Our method tackles the VOS task in an inherently different way than any of the previous papers in the literature. However, we adopt ideas presented in all of the above papers such as the use of ReID embedding vectors, optical flow proposal warping and fine-tuning on a large set of images augmented from the first frame.

%------------------------------------------------------------------------
\section{Approach}
\begin{figure}[t!]
  \centering
    \includegraphics[width=10cm]{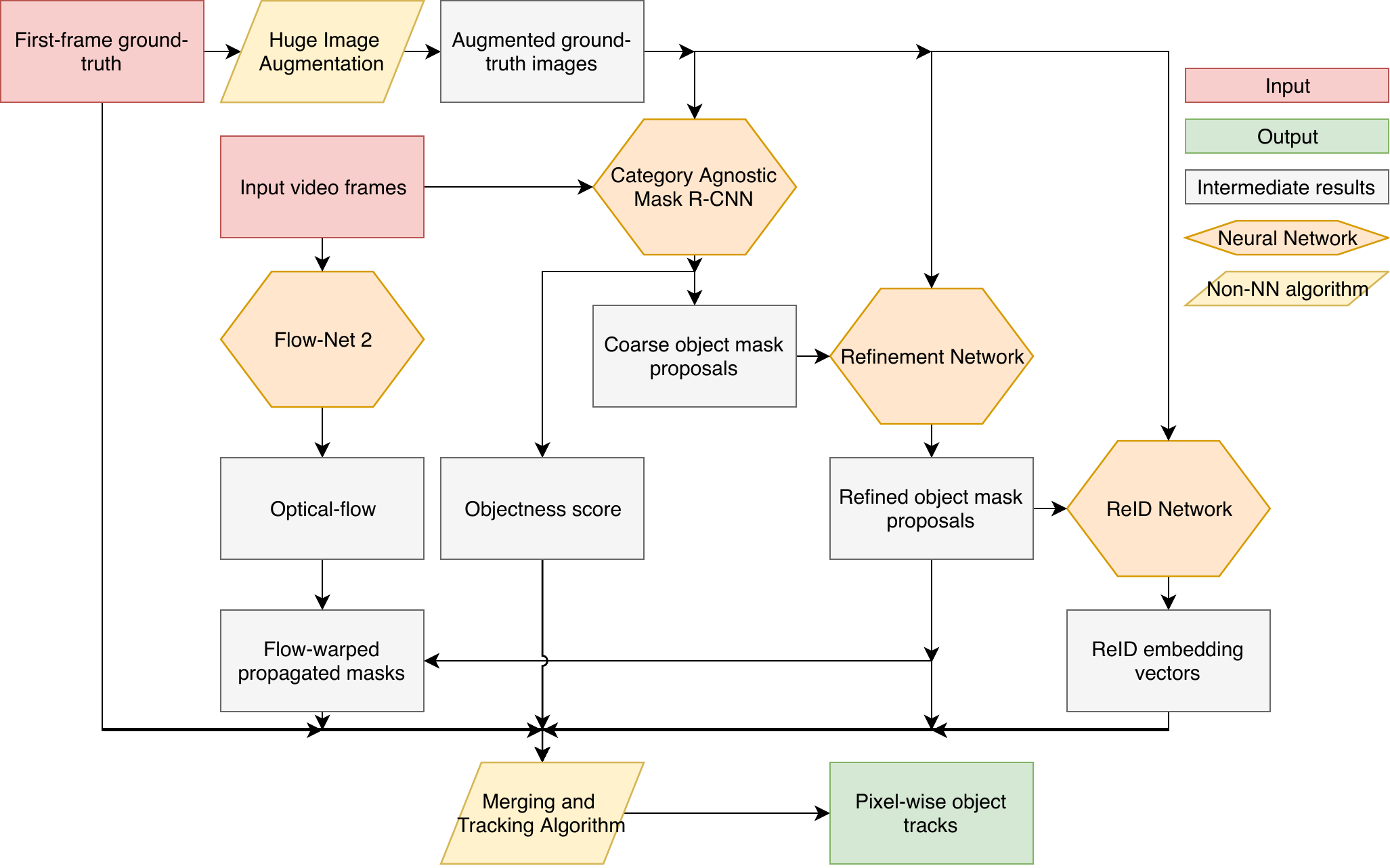}
  \caption{Diagram showing the components of PReMVOS and their relationships.}
  \label{fig:diag}
\end{figure}

We propose PReMVOS as a novel approach for addressing the VOS task. This approach is designed to produce more accurate and temporally consistent pixel masks across a video, especially in the challenging multi-object VOS task. Instead of predicting object masks directly on the video pixels, as done in \cite{caelles2017one}, \cite{maninis2017video}, \cite{voigtlaender17DAVIS} and \cite{voigtlaender17BMVC}, a key idea of our approach is to instead detect regions of interest as coarse object proposals using an object detection network, and to then predict accurate masks only on the cropped and resized bounding boxes.
We also present a new proposal merging algorithm in order to predict more temporally consistent pixel masks. The methods presented in \cite{Perazzi2017CVPR},\cite{li2017video},\cite{khoreva2017lucid},\cite{li2018video} create temporally consistent proposals by generating their proposals directly from the previous frame's proposals warped using optical flow into the current frame. Instead, our method generates proposals independently for each frame and then selects and links these proposals using a number of cues such as optical flow based proposal warping, ReID embeddings and objectness scores, as well as taking into account the presence of other objects in the multi-object VOS scenario.
This novel paradigm for solving the VOS task allows us to predict both more accurate and more temporally consistent pixel masks than all previous methods and achieves state-of-the-art results across all datasets. Figure \ref{fig:diag} shows an overview of each of the components of the PReMVOS algorithm and how these work together to solve the VOS task.

%-------------------------------------------------------------------------
\subsection{Image Augmentation}
For each video we generate a set of 2500 augmented images using the first-frame ground truth. We use the Lucid Data Dreaming method proposed in \cite{khoreva2017lucid} but only generate single images (not image pairs). This method removes the objects, automatically fills in the background, and then randomly transforms each object and the background before randomly reassembling the objects in the scene. Fine-tuning on this set of augmented images allows us to adapt our networks directly to the target video domain.

%-------------------------------------------------------------------------
\subsection{Proposal Generation}
We generate coarse object proposals using a Mask R-CNN \cite{he2017mask} network implementation by \cite{wu2016tensorpack} with a ResNet101 \cite{he2016deep} backbone. We adjust this network to be category agnostic by replacing the N classes with just one class by mapping all classes to a single foreground class for detecting generic objects. We train this network starting from pre-trained ImageNet \cite{Deng09CVPR} weights on both the COCO \cite{lin2014microsoft} and Mapillary \cite{neuhold2017mapillary} datasets jointly.  We then fine-tune a separate version of this network for each video for three epochs of the 2500 augmented images. This network generates coarse mask proposals, bounding boxes, and objectness scores for each image in the video sequence. We extract proposals with a score greater than 0.05 and also perform non-maximum suppression removing proposals which have an IoU of 66\% or greater with a proposal with a higher score.

%-------------------------------------------------------------------------
\subsection{Proposal Refinement}
The Proposal-Refinement Network is a fully convolutional network inspired by \cite{xu17BMVC} and based on the DeepLabv3+ \cite{chen2018encoder} architecture. This network takes as input a $385 \times 385$ image patch that has been cropped and resized from an approximate bounding box around an object of interest. A 50 pixel (in the original image) margin is first added to the bounding box in all directions. We add a fourth channel to the input image which encodes the original bounding box as a pixel mask to the input image. Starting from weights pre-trained on ImageNet \cite{Deng09CVPR}, COCO \cite{lin2014microsoft}, and PASCAL \cite{Everingham10IJCV}, we train this network on the Mapillary \cite{neuhold2017mapillary} dataset using random flipping, random gamma augmentations and random bounding box jitter \cite{xu17BMVC} up to 5\% in each dimension, to produce an accurate object segmentation, given an object's bounding box. We then fine-tune a separate version of this network for five epochs for each video on the 2500 augmented images. We then use this network to generate accurate pixel mask proposals for each of the previously generated coarse proposals, by only taking the bounding box of these proposals as input into the Refinement network and discarding the coarse mask itself.

%-------------------------------------------------------------------------
\subsection{Mask Propagation using Optical Flow}
As part of our proposal merging algorithm we use the optical flow between successive image pairs to warp a proposed mask into the next frame, to calculate the temporal consistency between two mask proposals. We calculate the Optical Flow using FlowNet 2.0 \cite{ilg2017flownet}.

%-------------------------------------------------------------------------
\subsection{ReID Embedding Vectors}
We further use a triplet-loss based ReID embedding network to calculate a ReID embedding vector for each mask proposal. We use the feature embedding network proposed in \cite{ovsep2017large}.
This is based on a wide ResNet variant \cite{wu2016wider} pre-trained on ImageNet \cite{Deng09CVPR} and then trained on the COCO dataset \cite{lin2014microsoft} using cropped bounding boxes resized to $128 \times 128$ pixels. This uses a triplet loss to learn an embedding space in which crops of different classes are separated and crops of the same class are grouped together. It is trained using the batch-hard loss with a soft-plus margin proposed in \cite{HermansBeyer2017Arxiv}.
We then fine-tune this network using the crops of each object from the generated 2500 augmented images for each of the 90 video sequences (242 objects) in the DAVIS 2017 \texttt{val}, \texttt{test-dev} and \texttt{test-challenge} datasets combined in order to have both enough positive and negative examples to train a network with a triplet-based loss. This network generates a ReID vector which differentiates all of the objects in these datasets from one other, which is used to compare the visual similarity of our generated object proposals and the first-frame ground truth object masks.

\begin{figure}[t!]
  \centering
    \includegraphics[width=10cm]{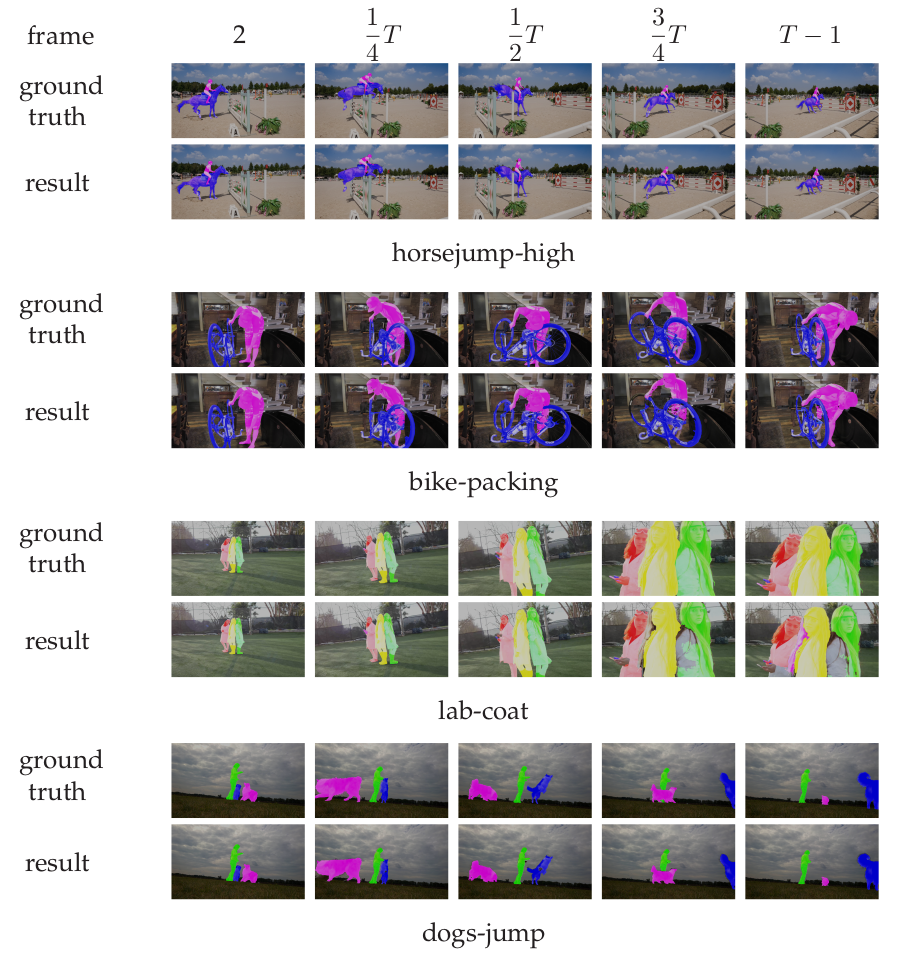}
  \caption{Qualitative results of PReMVOS on the DAVIS \texttt{2017 val dataset}.}
  \label{fig:viz}
\end{figure}

%-------------------------------------------------------------------------
\subsection{Proposal Merging}
Our proposal merging algorithm works in a greedy manner. Starting from the ground truth object masks in the first-frame, it builds tracks for each frame by scoring each of the proposals on their likeliness to belong to a particular object track. We have exactly one track for each ground truth object and we make hard decisions for which proposals to add to each track each timestep in a greedy manner. The proposal with the highest track score is added to each track. This track score is calculated as an affine combination of five separate sub-scores, each with values between 0 and 1. In the following, taking the complement of a score means subtracting it from 1.

The first sub-score is the \textit{Objectness} score. The objectness score $s_{obj,t,i,j}$ for the $j$-th track of the $i$-th proposal $c_{t,i}$ at time $t$ is given by
\begin{equation}
s_{obj,t,i,j}(c_{t,i}) = MaskObj(c_{t,i}),
\end{equation}
where $MaskObj(\cdot)$ denotes the confidence value provided by the Proposal Generation network.

The second score is a \textit{ReID} score, calculated using the Euclidean distance between the first-frame ground truth ReID embedding vector $r(f_j)$ and the ReID embedding vector $r(c_{t,i})$ of the current mask proposal, where $r(\cdot)$ denotes applying the ReID network, $\lVert \cdot \rVert$ denotes the L2 norm, and $f_j$ is the bounding box of the $j$-th ground truth object in the first frame. This distance is then normalized by dividing it by the maximum distance for all proposals in a video from the ground truth embedding vector of interest. The complement is then taken to convert from a distance into a similarity score.
\begin{equation}
s_{reid,t,i,j}(c_{t,i}, f_j)= 1 - \frac{\lVert r(c_{t,i}) - r(f_j) \rVert}{\max_{\tilde{t}, \tilde{i}} \lVert r(c_{\tilde{t},\tilde{i}}) - r(f_j) \rVert}
\end{equation}

The third score is a \textit{Mask Propagation} score. This is calculated for each possible object track as the IoU between the current mask proposal and the warped proposal that was decided for in the previous time-step for this object track, warped into the current time-step using the optical flow:
\begin{equation}
s_{maskprop, t, i, j}(c_{t,i}, p_{t-1,j}) = IoU(c_{t,i}, warp(p_{t-1,j})),
\end{equation}
where $p_{t-1,j}$ is is the previously selected proposal for timestep $t-1$ for object track $j$ and $warp(\cdot)$ applies optical flow mask warping from frame $t-1$ to $t$.

The fourth score is an \textit{Inverse ReID} score. This is calculated as the complement of the maximum \textit{ReID} score for the current mask proposal and all other object tracks $k$ except the object track of interest $j$:
\begin{equation}
s_{inv\_reid,t,i,j} = 1 - \max_{k \neq j}(s_{reid,t,i,k}).
\end{equation}
The fifth score is an \textit{Inverse Mask Propagation} score. This is calculated as the complement of the maximum \textit{Mask Propagation IoU} score for the current mask proposal and all other object tracks $k$ except the object track of interest $j$:
\begin{equation}
s_{inv\_maskprop,t,i,j} = 1 - \max_{k \neq j}(s_{maskprop,t,i,k}).
\end{equation}

All five scores are combined together by
\begin{equation} s_{comb, t, i, j} = \sum_{q \in \{objectness, reid, maskprop, inv\_reid, inv\_maskprop\}} \alpha_q s_{q,t,i,j},
\end{equation}
where $\sum_q \alpha_q = 1$ and all $\alpha_q \geq 0$. The greedy decisions are then made by 
$p_{t,j}=c_{t,k_j}$, where 
\begin{equation}
k_j=\argmax_i{s_{comb, t, i, j}}.
\end{equation}

In cases where the selected proposals for the different objects within one time-step overlap, we assign the overlapping pixels to the proposal with the highest combined track score.
We present results with both an equal weighting for each of the five sub-score components and where the weights are tuned using random-search hyper-parameter optimisation evaluated against the DAVIS 2017 \texttt{val} set. The values of these optimised weights are shown in Table \ref{table:merge}. We ran this optimisation for 25000 random parameter values. For the results on the 2018 DAVIS Challenge we also present results using an ensemble of the results using the top 11 sets of parameter values, using a simple pixel-wise majority vote to ensemble the results.

%------------------------------------------------------------------------
% MAIN RESULTS TABLE
\newcommand{\mc}[1]{\shortstack{#1}}
\begin{table*}[t!]
\footnotesize
\newcolumntype{Y}{>{\centering\arraybackslash\hsize=0.09\hsize}X}
\newcolumntype{Z}{>{\centering\arraybackslash\hsize=0.05\hsize}X}
\newcolumntype{W}{>{\centering\arraybackslash\hsize=0.11\hsize}X}
\newcolumntype{V}{>{\centering\arraybackslash\hsize=0.12\hsize}X}
\newcolumntype{S}{>{\centering\arraybackslash\hsize=0.08\hsize}X}
\newcommand{\mmr}{\arrayrulecolor{lightgray}\cmidrule[0.25pt]{2-12}\arrayrulecolor{black}}

\setlength{\tabcolsep}{0pt}
\begin{center}
%\begin{tabularx}{\textwidth}{|cX|cX|cX|cX|cX|cX|cX|cX|cX|cX|cX|cX|}
%\begin{tabularx}{\textwidth}{Y>{\centering\arraybackslash}p{1.1cm}YYYYYYYYYYYY>{\centering\arraybackslash}p{1.2cm}WW}
\begin{tabularx}{\textwidth}{ZSSYYYYYVWWY}
\toprule[1pt]
\multicolumn{3}{c}{} & \multirow{2}{*}{Ours} & DyeNet & MRF & Lucid & ReID & OSVOS-S & OnAVOS & OSVOS \\

\multicolumn{3}{c}{} & & \cite{li2018video} & \cite{bao2018cnn} & \cite{khoreva2017lucid} & \cite{li2017video} & \cite{maninis2017video} & \cite{voigtlaender17DAVIS}\cite{voigtlaender17BMVC} & \cite{caelles2017one} \\

%%%%%%%%%%%%%%%%%%%%%%%%%%%%%%%%%%%%%%%%%%%%%%%%%%%%%%%%%%%%%%%%%%%%%%%%%%%%%%%%%%

\midrule[0.5pt]\multirow{7}{*}{\mc{17 \\ T-D}}  &  $\mathcal{J}$\&$\mathcal{F}$ & Mean & \textbf{71.6} & 68.2 & 67.5 & 66.6 & 66.1 & 57.5 & 56.5 & 50.9\\
%\hline
\mmr
 & \multirow{3}{*}{$\mathcal{J}$} & Mean & \textbf{67.5} & 65.8 & 64.5 & 63.4 & 64.4 & 52.9 & 52.4 & 47.0\\
 & & Recall & \textbf{76.8} & - & - & 73.9 & - & 60.2 & - & 52.1\\
 & & Decay & 21.7 & - & - & 19.5 & - & 24.1 & - & \textbf{19.2}\\
\mmr
 & \multirow{3}{*}{$\mathcal{F}$} & Mean & \textbf{75.7} & 70.5 & 70.5 & 69.9 & 67.8 & 62.1 & 59.6 & 54.8\\
 & & Recall & \textbf{84.3} & - & - & 80.1 & - & 70.5 & - & 59.7\\
 & & Decay & 20.6 & - & - & \textbf{19.4} & - & 21.9 & - & 19.8\\

%%%%%%%%%%%%%%%%%%%%%%%%%%%%%%%%%%%%%%%%%%%%%%%%%%%%%%%%%%%%%%%%%%%%%%%%%%%%%%%%%%

\midrule[0.5pt]\multirow{7}{*}{\mc{17 \\ Val}} & $\mathcal{J}$\&$\mathcal{F}$ & Mean & \textbf{77.8} & 74.1 & 70.7 & - & - & 68.0 & 67.9 & 60.3\\
\mmr
 & \multirow{3}{*}{$\mathcal{J}$} & Mean & \textbf{73.9} & - & 67.2 & - & - & 64.7 & 64.5 & 56.6\\
 & & Recall & \textbf{83.1} & - & - & - & - & 74.2 & - & 63.8\\
 & & Decay & 16.2 & - & - & - & - & \textbf{15.1} & - & 26.1\\
\mmr
 & \multirow{3}{*}{$\mathcal{F}$} & Mean & \textbf{81.7} & - & 74.2 & - & - & 71.3 & 71.2 & 63.9\\
 & & Recall & \textbf{88.9} & - & - & - & - & 80.7 & - & 73.8\\
 & & Decay & 19.5 & - & - & - & - & \textbf{18.5} & - & 27.0\\

%%%%%%%%%%%%%%%%%%%%%%%%%%%%%%%%%%%%%%%%%%%%%%%%%%%%%%%%%%%%%%%%%%%%%%%%%%%%%%%%%%

\midrule[0.5pt]
\multirow{8}{*}{\mc{16 \\ Val}} & $\mathcal{J}$\&$\mathcal{F}$ & Mean & \textbf{86.8} & - & - & - & - & 86.5 & 85.5 & 80.2\\
\mmr
 & \multirow{3}{*}{$\mathcal{J}$} & Mean & 84.9 & \textbf{86.2} & 84.2 & - & - & 85.6 & 86.1 & 79.8\\
 & & Recall & 96.1 & - & - & - & - & \textbf{96.8} & 96.1 & 93.6\\
 & & Decay & 8.8 & - & - & - & - & 5.5 & \textbf{5.2} & 14.9\\
\mmr
 & \multirow{3}{*}{$\mathcal{F}$} & Mean & \textbf{88.6} & - & - & - & - & 87.5 & 84.9 & 80.6\\
 & & Recall & 94.7 & - & - & - & - & \textbf{95.9} & 89.7 & 92.6\\
 & & Decay & 9.8 & - & - & - & - & 8.2 & \textbf{5.8} & 15.0\\
\mmr
 & $\mathcal{T}$ & Mean & 36.4 & - & - & - & - & 21.7 & \textbf{19.0} & 37.8\\

%%%%%%%%%%%%%%%%%%%%%%%%%%%%%%%%%%%%%%%%%%%%%%%%%%%%%%%%%%%%%%%%%%%%%%%%%%%%%%%%%%

\bottomrule[1pt]
\end{tabularx}
\end{center}
\caption{Our results and other state-of-the-art results on the three DAVIS datasets: the 2017 \texttt{test-dev} set (17 T-D), the 2017 \texttt{val} set (17 Val), and the 2016 {val} set (16 Val). On the 17 Val and 16 Val datasets we use the naive merging component weights, whereas on the 17 T-D dataset we use the weights optimised using the 17 Val set.}
\label{table:results}
\end{table*}

%%%%%%%%%%%%%%%%%%%%%%%%%%%%%%%%%%%%%%%%%%%%%%%%%%%%%%%%%%%%%%%%%%%%%%%%%%%%%%%%%%%%%%%%%%%%%%%%%%%%%%%%%%%%%%%%%%%%%%%%%%%%%%%%%%%%%%%%%%%%%%%%%%%%%%%%%%%%%%%%%%%%%%%%%%%%%%%%%%%%%%%%%%%%%%%%%%%%%%%%%%%%%%%%%%%%%%%%
% 2018 challenge table
\begin{table*}[t!]
\footnotesize
\newcolumntype{Y}{>{\centering\arraybackslash\hsize=0.15\hsize}X}
\newcolumntype{Z}{>{\centering\arraybackslash\hsize=0.15\hsize}X}
\newcolumntype{W}{>{\centering\arraybackslash\hsize=0.25\hsize}X}
\newcolumntype{S}{>{\centering\arraybackslash\hsize=0.28\hsize}X}
\newcolumntype{V}{>{\centering\arraybackslash\hsize=0.01\hsize}X}
\newcommand{\mmr}{\arrayrulecolor{lightgray}\cmidrule[0.25pt]{1-10}\arrayrulecolor{black}}

\setlength{\tabcolsep}{0pt}
\begin{center}
%\begin{tabularx}{\textwidth}{|cX|cX|cX|cX|cX|cX|cX|cX|cX|cX|cX|cX|}
%\begin{tabularx}{\textwidth}{YYYYYYYYYYYY}
\begin{tabularx}{\textwidth}{ZZYYYWWYSV}
%\begin{tabularx}{\textwidth}{XXXXXXXXXXXXXXXXX}
\toprule[1pt]
\multicolumn{2}{c}{} & Ours & \multirow{2}{*}{Ours} & DyeNet & ClassAgno. & OnlineGen. & Lucid & ContextBased \\

\multicolumn{2}{c}{} & (Ens) & & \cite{DAVIS2018-Semi-Supervised-2nd} & VOS \cite{DAVIS2018-Semi-Supervised-3rd} & VOS \cite{DAVIS2018-Semi-Supervised-4th} & \cite{DAVIS2018-Semi-Supervised-5th} & VOS \cite{DAVIS2018-Semi-Supervised-6th} &\\

%%%%%%%%%%%%%%%%%%%%%%%%%%%%%%%%%%%%%%%%%%%%%%%%%%%%%%%%%%%%%%%%%%%%%%%%%%%%%%%%%%%%

\midrule[0.5pt]
$\mathcal{J}$\&$\mathcal{F}$ & Mean & \textbf{74.7} & 71.8 & 73.8 & 69.7 & 69.5 & 67.8 & 66.3\\
\mmr
\multirow{3}{*}{$\mathcal{J}$} & Mean & 71.0 & 67.9 & \textbf{71.9} & 66.9 & 67.5 & 65.1 & 64.1\\
 & Recall & \textbf{79.5} & 75.9 & 79.4 & 74.1 & 77.0 & 72.5 & 75.0\\
 & Decay & 19.0 & 23.2 & 19.8 & 23.1 & 15.0 & 27.7 & \textbf{11.7}\\
\mmr
\multirow{3}{*}{$\mathcal{F}$} & Mean & \textbf{78.4} & 75.6 & 75.8 & 72.5 & 71.5 & 70.6 & 68.6\\
 & Recall & \textbf{86.7} & 82.9 & 83.0 & 80.3 & 82.2 & 79.8 & 80.7\\
 & Decay & 20.8 & 24.7 & 20.3 & 25.9 & 18.5 & 30.2 & \textbf{13.5}\\
\bottomrule[1pt]
%\end{tabularx}
\end{tabularx}
\end{center}
\caption{Our results (with and without ensembling) on the DAVIS \texttt{test-challenge} dataset compared with the top five other competitors in the 2018 DAVIS Challenge.}
\label{table:challenge}
\end{table*}

%------------------------------------------------------------------------
\section{Experiments}
We evaluate our algorithm on the set of DAVIS \cite{Caelles_arXiv_2018},\cite{Pont-Tuset_arXiv_2017},\cite{Perazzi2016} datasets and benchmarks. Table \ref{table:results} shows our results on the three DAVIS benchmarks. The DAVIS 2017 \texttt{test-dev} and \texttt{val} datasets contain multiple objects per video sequence, whereas the DAVIS 2016 \texttt{val} dataset contains a single object per sequence. The metrics of interest are the $\mathcal{J}$ score, calculated as the average IoU between the proposed masks and the ground truth mask, and the $\mathcal{F}$ score, calculated as an average boundary similarity measure between the boundary of the proposed masks and the ground truth masks. For more details on these metrics see \cite{Perazzi2016}.

On all of the datasets our method gives results better than all other state-of-the-art methods for both the $\mathcal{F}$ metric and the mean of the $\mathcal{J}$ and $\mathcal{F}$ score. We also produce either the best, or comparable to the best, results on the $\mathcal{J}$ metric for each dataset. These results show that the novel proposed VOS paradigm performs better than the current VOS paradigms in predicting both accurate and temporally consistent mask proposals.

Table \ref{table:challenge} shows our results both with and without ensembling on the DAVIS 2017/2018 \texttt{test-challenge} dataset evaluated during the 2018 DAVIS Challenge compared to the top six other competing methods. Our method gives the best results and gets first place in the 2018 DAVIS Challenge. 

Figure \ref{fig:viz} shows qualitative results of our method on four video sequences from the \texttt{2017 val} dataset. These results show that our method produces both accurate and temporally consistent results across the video sequences.

%%%%%%%%%%%%%%%%%%%%%%%%%%%%%%%%%%%%%%%%%%%%%%%%%%%%%%%%%%%%%%%%%%%%%%%%

\begin{table*}[t!]
\footnotesize
\newcolumntype{Y}{>{\centering\arraybackslash\hsize=0.21\hsize}X}
\newcolumntype{Z}{>{\centering\arraybackslash\hsize=0.11\hsize}X}
\newcolumntype{S}{>{\centering\arraybackslash\hsize=0.30\hsize}X}
\newcommand{\mmr}{\arrayrulecolor{lightgray}\cmidrule[0.25pt]{1-5}\arrayrulecolor{black}}
\setlength{\tabcolsep}{0pt}
\begin{center}
\begin{tabularx}{\textwidth}{SYYYZ}
\toprule[1pt]

& $\mathcal{J}$ mean &  $\mathcal{F}$ mean & $\mathcal{J}$\&$\mathcal{F}$ mean &\\

\midrule[0.5pt]
Without Refinement & 71.2 & 77.3 & 74.2 &\\
\mmr
With Refinement & \textbf{77.1} & \textbf{85.2} & \textbf{81.2} & \\
\mmr
Boost & 5.9 & 7.9 & 7.0 & \\

\bottomrule[1pt]
\end{tabularx}
\end{center}
\caption{Quantitative results of an ablation study on the \texttt{2017 val} dataset showing the effect of the Refinement Network on the accuracy of generated mask proposals. Presented results are calculated using \textit{oracle merging} (see Section \ref{section:ref}).}
\label{table:ref}
\end{table*}

%%%%%%%%%%%%%%%%%%%%%%%%%%%%%%%%%%%%%%%%%%%%%%%%%%%%%%%%%%%%%%%%%%%%%%%%%%%%%%%%%%%%%%%%%%%%%%%%%%%%%%%%%%%%%%%%%%%%%%%%%%%%%%%%%%%%%%%%%%%%%%%%

\begin{figure}[t!]
  \centering
    \includegraphics[width=10cm]{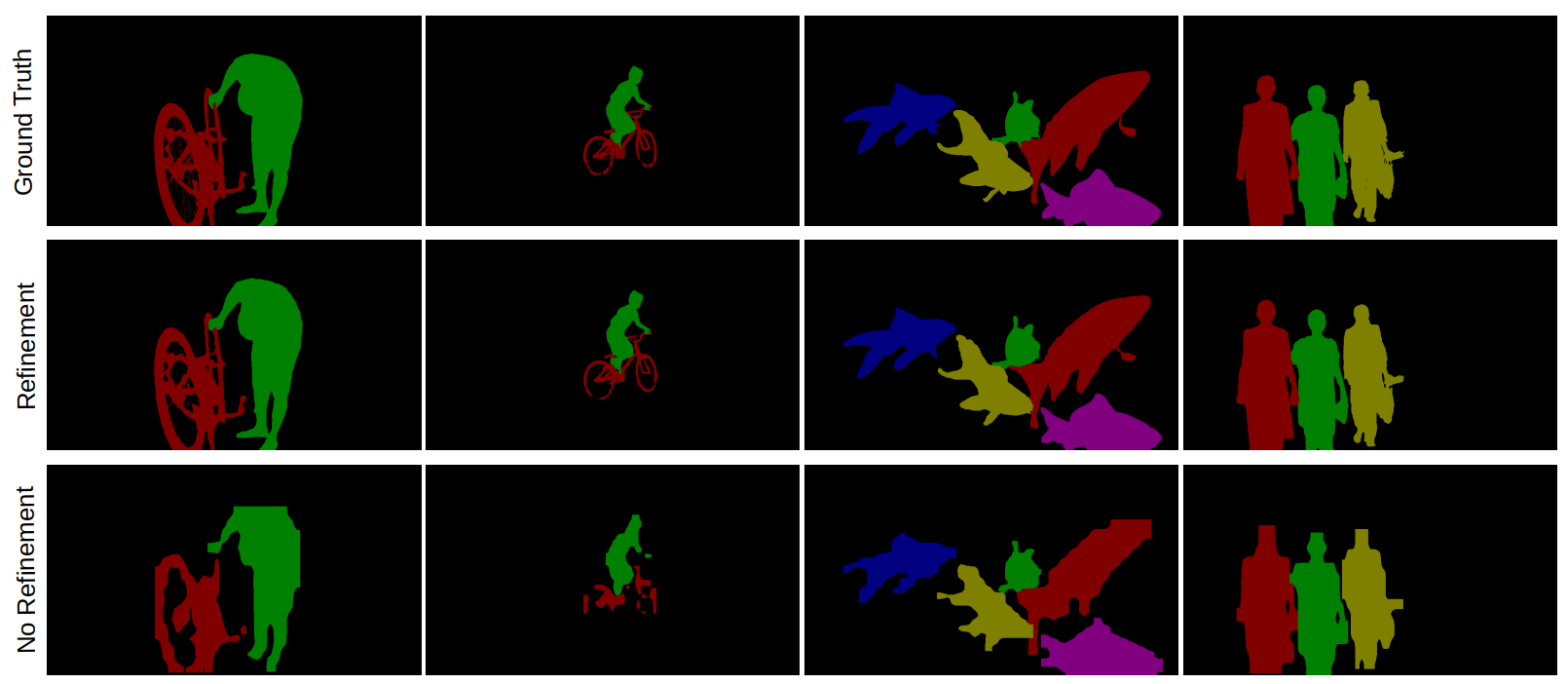}
  \caption{Qualitative results showing the effect of the Refinement Network on the mask proposal accuracy. Results are calculated using \textit{oracle merging} (Section \ref{section:ref}).}
  \label{fig:ref}
\end{figure}

%%%%%%%%%%%%%%%%%%%%%%%%%%%%%%%%%%%%%%%%%%%%%%%%%%%%%%%%%%%%%%%%%%%%%%%%

%-------------------------------------------------------------------------
\subsection{Proposal Refinement}
\label{section:ref}
We perform an ablation study to investigate the effect of the Refinement network on the accuracy of the mask proposals. We compare the coarse proposals generated from a state-of-the-art instance segmentation method, Mask R-CNN \cite{he2017mask}, to the refined proposals generated by feeding the bounding boxes of these coarse proposals into our Refinement Network. In order to evaluate these results we calculate the best proposal for each object in each frame which maximizes the IoU score with the ground-truth mask. We then evaluate using the standard DAVIS evaluation metrics using these merged proposals. This \textit{oracle merging} algorithm allows us to evaluate the accuracy of our proposal generation separately from our merging algorithm.

Table \ref{table:ref} shows the quantitative IoU ($\mathcal{J}$) and boundary measure ($\mathcal{F}$) improvement for our refined proposals over the Mask R-CNN proposals. Our method results in a $5.9\%$ improvement in IoU and a $7.9\%$ improvement in boundary measure over just using the Mask R-CNN proposals.

Figure \ref{fig:ref} also visualizes the qualitative results of the improved accuracy of the refined masks over the generated Mask R-CNN masks. In all examples the refined proposals match the ground truth masks more closely than the coarse proposals and capture the boundary contours at a much higher fidelity. This is due to the refinement network extracting deep features only over the area of interest for each object, and not over the whole image as is done in the case for Mask R-CNN. The Refinement Network is also able to recover parts of objects that were completely lost in the coarse proposals, for example in the two examples of bicycles shown. This is because the Refinement Network takes as input only the bounding box of the coarse mask proposal, not the coarse mask itself, and relies only on it's trained knowledge of what is inherently an \textit{object} in order to generate the refined segmentation masks. Also when the bounding box does not accurately cover the whole object, as is the case with the bicycle in the second column of Figure \ref{fig:ref}, the refinement network is still able to recover the accurate mask as it takes into account a 50 pixel margin around the coarse mask proposal bounding box.

%%%%%%%%%%%%%%%%%%%%%%%%%%%%%%%%%%%%%%%%%%%%%%%%%%%%%%%%%%%%%%%%%%%%%%%%%%%%%%%%%%%%%%%%%%%%%%%%%%%%%%%%%%%%%%%%%%%%%%%%%%%%%%%%%%%%%%%%%%%%%%%%

\begin{table*}[]
\footnotesize
\newcolumntype{Y}{>{\centering\arraybackslash\hsize=0.21\hsize}X}
\newcolumntype{Z}{>{\centering\arraybackslash\hsize=0.24\hsize}X}
\newcolumntype{S}{>{\centering\arraybackslash\hsize=0.15\hsize}X}
\newcommand{\mmr}{\arrayrulecolor{lightgray}\cmidrule[0.25pt]{1-7}\arrayrulecolor{black}}
\newcommand{\mr}{\arrayrulecolor{lightgray}\cmidrule[0.25pt]{2-6}\arrayrulecolor{black}}
\setlength{\tabcolsep}{0pt}
\begin{center}
\begin{tabularx}{\textwidth}{SYYYYZY}
\toprule[1pt]
Num. & \multicolumn{5}{c}{Merging Sub-Score Components} & $\mathcal{J}$\&$\mathcal{F}$\\
\mr
Comp. & Objectness & ReID & InvReID & MaskProp & InvMaskProp & Mean\\
\midrule[0.5pt]

0 & \multicolumn{5}{c}{Oracle merging} & \textbf{81.2}\\

\mmr
5(opt.) & 19\% & 18\% & 14\% & 22\% & 27\% & \textbf{78.2}\\
\mmr

5 & \checkmark & \checkmark & \checkmark & \checkmark & \checkmark & \textbf{77.8}\\
\mmr

\multirow{5}{*}{4} & \checkmark & \checkmark & \checkmark & \checkmark & - & 76.7\\
& \checkmark & \checkmark & \checkmark & - & \checkmark & \textit{75.5}\\
& \checkmark & \checkmark & - & \checkmark & \checkmark & \textbf{76.9}\\
&\checkmark & - & \checkmark & \checkmark & \checkmark & 76.3\\
& - & \checkmark & \checkmark & \checkmark & \checkmark & 75.9\\

\mmr

\multirow{10}{*}{3} & \checkmark & \checkmark & \checkmark & - & - & 74.2\\
& \checkmark & \checkmark & - & \checkmark & - & 75.0\\
& \checkmark & \checkmark & - & - & \checkmark &  74.2\\
& \checkmark & - & \checkmark & \checkmark & - & 73.5\\
& \checkmark & - & \checkmark & - & \checkmark & 69.6\\
& \checkmark & - & - & \checkmark  & \checkmark & 71.1\\
& - & \checkmark & \checkmark & \checkmark & - & 75.8\\
& - & \checkmark & \checkmark & - & \checkmark & \textit{69.3}\\
& - & \checkmark & - & \checkmark  & \checkmark & \textbf{75.9}\\
& - & - & \checkmark & \checkmark  & \checkmark & 74.3\\

\mmr

\multirow{10}{*}{2} & \checkmark & \checkmark & - & - & - & 72.7\\
& \checkmark & - &\checkmark & - & - & 64.7\\
& \checkmark & - & - & \checkmark & - &  69.1\\
& \checkmark & - & - & - & \checkmark & 57.9\\
& - & \checkmark & \checkmark & - & - & 68.7\\
& - & \checkmark & - & \checkmark & - & \textbf{74.3}\\
& - & \checkmark & - & - & \checkmark & 68.8\\
& - & - & \checkmark & \checkmark & - & 74.0\\
& - & - & \checkmark & - & \checkmark & \textit{47.3}\\
& - & - & - & \checkmark & \checkmark & 73.6\\

\mmr

\multirow{5}{*}{1} & \checkmark & - & - & - & - & \textit{29.5}\\
 & - & \checkmark & - & - & - & 67.4\\
 & - & - & \checkmark & - & - & 44.3\\
 & - & - & - & \checkmark & - & \textbf{72.8}\\
 & - & - & - & - & \checkmark & 34.4\\

\bottomrule[1pt]
\end{tabularx}
\end{center}
\caption{Results of an ablation study on the DAVIS \texttt{2017 val} dataset showing the effect of each of the merging algorithm sub-score components on the accuracy of the merging algorithm. The \textit{oracle merging} result indicates an upper bound for the merging algorithm performance (Section \ref{section:ref}). The components given as percentages indicate the optimal component weights calculated using hyper-parameter optimisation on the \texttt{2017 val} set. The components given by a checkmark (\checkmark) have equal weights for each checked components. For each group of results with the same number of components the best result is expressed in \textbf{bold} and the worst in \textit{italics}.}
\label{table:merge}
\end{table*}

%-------------------------------------------------------------------------
\subsection{Proposal Merging}
We perform a further ablation study to investigate the effect of each of the merging algorithm sub-score components on the accuracy of the merging algorithm. Table \ref{table:merge} shows the results of this ablation study on the DAVIS \texttt{2017 val} dataset.

We first present an upper bound baseline result for our merging algorithm. This is calculated by choosing the best proposal for each object in each frame which maximizes the IoU score with the ground-truth mask. This \textit{oracle merging} method gives a $81.2$ $\mathcal{J}$\&$\mathcal{F}$ mean score.

We then present the results of our merging algorithm where the weights for each of the five sub-score components were optimised using random-search hyper-parameter optimisation evaluated against the DAVIS \texttt{2017 val} set. This optimisation was done by evaluating 25000 random affine combinations of the 5 component weights and selecting the set of weights that resulted in the best IoU score. This result gives another upper bound of $78.2$ $\mathcal{J}$\&$\mathcal{F}$ mean score, as this was evaluated on the same dataset as the weight values were optimised on. Our greedy selection algorithm based on a combination of the 5 sub-component scores is able to reach a $\mathcal{J}$\&$\mathcal{F}$ mean score that is only $3\%$ lower than the hypothetical maximum, showing that these carefully selected 5 sub-score components are sufficient to generate accurate and consistent object tracks, even in difficult cases such as multiple similar objects and large occlusions. These opimised weights used on the \texttt{2017 test-dev} dataset presented in Table \ref{table:results}, all other dataset results in Table \ref{table:results} and Table \ref{table:merge} use equal weights for the five sub-components.

The naive combination with all 5 sub-score components has a $\mathcal{J}$\&$\mathcal{F}$ mean score of 77.8 which is only $0.4$ below that with optimised weights. This indicates that the merging algorithm is relatively robust to the exact weights and that what is more important is the presence of all five components.

The \textit{Objectness} sub-score separates well-defined mask proposals with accurate boundary contours from proposals with boundaries that are less likely to model a consistent object. It is also able to distinguish objects of interest given in the first frame from other objects in the scene, as this score was trained in the Mask R-CNN fine-tuning process to identify these objects and ignore the others. However, this sub-score component is unable to distinguish between different objects of interest if more than one is present in a video sequence.

The \textit{ReID} sub-score is used to distinguish between objects that look visually different from each other. This works well to separate objects such as bikes and people from each other in the same video sequence, but does not work as well on sequences with multiple similar looking objects.

The \textit{MaskProp} sub-score is used for temporal consistency. This can distinguish well between very similar objects if they are separated in the spatial domain of the image. However, this sub-score cannot deal with cases where objects heavily occlude each other, or completely disappear before later returning.

The \textit{InvReID} and \textit{InvMaskProp} sub-scores are used to force the selected mask proposals in each frame to be as distinguishable from each other as possible. Just using the other 3 components often results in failure cases where the same or very similar mask proposals are chosen for different objects. This occurs when similar looking objects overlap or when one object disappears. These two sub-components work by distinguishing proposals that are visually and temporally inconsistent with other objects in the video sequence, resulting in a signal of consistency with the object of interest. These components can separate well between the different objects of interest in a video sequence, but they are unable to separate these objects from possible background objects.

The results in Table \ref{table:merge} show that all five sub-scores are important for accurate proposal merging, as removing one of these components results in a loss of accuracy between $0.9$, when only removing the \textit{InvReID} sub-score, to $2.3$, when removing the \textit{MaskProp} sub-score. Without both the \textit{InvReID} and \textit{InvMaskProp}, the $\mathcal{J}$\&$\mathcal{F}$ mean score decreases by $1.9$ points, showing that these sub-scores that were introduced to promote spatial separation of the chosen proposals are an integral part of the merging algorithm. Only using these two sub-components, however, results in a score decrease of $30.5$ points. Removing the two main key components, the \textit{ReID} and \textit{MaskProp} sub-scores, results in a loss of $8.2$ points, whereas only having these two components results in a loss of $3.5$ points. When used by itself, the \textit{MaskProp} sub-score is the strongest component, resulting in a loss of $5.0$ points.

\subsection{Runtime Evaluation}
\label{sec:runtime}
We perform a runtime evaluation of the different components of the PReMVOS algorithm. The complete PReMVOS algorithm with first-frame data augmentation and fine-tuning was designed in order to produce the most accurate results without regards for speed requirements. For further evaluation, we present in Table \ref{table:time} three versions of PReMVOS. The original version, a fast-finetuned version, and a version without any fine-tuning. The fast-finetuned version is fine-tuned for one third of the iterations as the original method, and instead of the slow Lucid Data Dreaming \cite{khoreva2017lucid} image augmentations, it uses simple rotations, translations, flipping and brightness augmentations. Only one set of weights are fine-tuned over the whole DAVIS val set rather than different weights for each video. A combination of the \textit{specific} proposals from the fine-tuned proposal network and the \textit{general} proposals from a proposal network that was not fine-tuned on the validation set first-frames are used. The not-finetuned version of PReMVOS just uses the \textit{general} proposals without any fine-tuning.
Figure \ref{fig:speed} compares the quality and runtime of PReMVOS against other methods in the literature. Across all of the presented runtime scales, our method compares to or exceeds all other state-of-the-art results.

%%%%%%%%%%%%%%%%%%%%%%%%%%%%%%%%%%%%%%%%%%%%%%%%%%%%%%%%%%%%%%%%%%%%%%%%

\begin{table*}[t!]
\footnotesize
%\newcolumntype{Y}{>{\centering\arraybackslash\hsize=0.21\hsize}X}
%\newcolumntype{Z}{>{\centering\arraybackslash\hsize=0.11\hsize}X}
%\newcolumntype{S}{>{\centering\arraybackslash\hsize=0.60\hsize}X}

\newcolumntype{Y}{>{\centering\arraybackslash\hsize=0.075\hsize}X}
\newcolumntype{Z}{>{\centering\arraybackslash\hsize=0.11\hsize}X}
\newcolumntype{S}{>{\centering\arraybackslash\hsize=0.18\hsize}X}

\newcommand{\mmr}{\arrayrulecolor{lightgray}\cmidrule[0.25pt]{1-14}\arrayrulecolor{black}}
\setlength{\tabcolsep}{0pt}
\begin{center}
%\begin{tabularx}{\textwidth}{SYYYZ}
%\begin{tabularx}{\textwidth}{SXXXXXXXXXXX}
\begin{tabularx}{\textwidth}{SYYYYYYYYYYYYY}
\toprule[1pt]

 & \rotatebox{90}{Augm. Gen.} &  \rotatebox{90}{Fine-tuning} & \rotatebox{90}{Prop. Gen.} &\rotatebox{90}{Prop. Refine.} & \rotatebox{90}{ReID} & \rotatebox{90}{Optic. Flow} & \rotatebox{90}{Warping} & \rotatebox{90}{Merging} & \rotatebox{90}{\textbf{Total}} &  \rotatebox{90}{Av. \# Prop.} & \rotatebox{90}{Mean $\mathcal{J}$\&$\mathcal{F}$} & \\

\midrule[0.5pt]
Original & 23.4 & 12.3 & 0.41 & 1.04 & 0.05 & 0.14 & 0.32 & 0.02 & 37.4 & 17.52 & 77.8\\
\mmr
Fast-finetuned & 0.02 & 3.9  & 0.26 & 0.45 & 0.03 & 0.14 & 0.20 & 0.02 & 5.02 & 9.28 & 73.7\\
\mmr
Not-finetuned & 0.00 & 0.00 & 0.14 & 0.33 & 0.02 & 0.14 & 0.16 & 0.02 & 0.81 & 6.87 & 65.7\\

\bottomrule[1pt]
\end{tabularx}
\end{center}
\caption{Runtime analysis of the different components of the PReMVOS algorithm. Times are in seconds per frame, averaged over the DAVIS 2017 val set. Augmentation Generation is run on 48 CPU cores, and Fine-tuning is done on 8 GPUs. Otherwise, everything is run sequentially on one GPU / CPU core.}
\label{table:time}
\end{table*}

%%%%%%%%%%%%%%%%%%%%%%%%%%%%%%%%%%%%%%%%%%%%%%%%%%%%%%%%%%%%%%%%%%%%%%%%%%%

\begin{figure}[t!]
\centering
\resizebox{\linewidth}{!}{\begin{tikzpicture}[/pgfplots/width=1\linewidth, /pgfplots/height=0.4\linewidth, /pgfplots/legend pos=south east]
    \begin{axis}[ymin=50,ymax=80,xmin=0.1,xmax=100,enlargelimits=false,
        xlabel=Time per frame (seconds),
        ylabel=Region and contour quality ($\mathcal{J}$ \& $\mathcal{F}$),
		font=\normalsize,
        grid=both,
		grid style=dotted,
        xlabel shift={-2pt},
        ylabel shift={-5pt},
        xmode=log,
        legend columns=1,
        %transpose legend,
        minor ytick={0,2.5,...,110},
        ytick={0,10,...,110},
	    yticklabels={0,10,20,30,40,50,60,70,80,90,100},
	    xticklabels={0.01,.1,1,10,100},
        legend pos= outer north east,
        legend cell align={left}
        ]

	\addplot[red,mark=star,only marks,line width=2, mark size=5.0] coordinates{(37.4,77.8)};
        \addlegendentry{\hphantom{i}Original (Ours)}
        %if we need it, here \label{...} can be added
        
        \addplot[red,mark=+,only marks,line width=2, mark size=5.0] coordinates{(5.02,73.7)};
        \addlegendentry{\hphantom{i}Fast-finetuned (Ours)}
       
        \addplot[red,mark=x,only marks,line width=2, mark size=5.0] coordinates{(0.81,65.7)};
        \addlegendentry{\hphantom{i}Not-finetuned (Ours)}
        
        \addplot[green,mark=o, mark size=3.5,only marks, line width=0.75] coordinates{(18,60.3)};
        \addlegendentry{\hphantom{i}OSVOS \cite{caelles2017one}}
      
        \addplot[black,mark=o,only marks,line width=0.75, mark size=3.5] coordinates{(26,	67.9)};
        \addlegendentry{\hphantom{i}OnAVOS \cite{voigtlaender17BMVC}}
        
        \addplot[olive,mark=o, mark size=3.5,only marks, line width=0.75] coordinates{(9,68.)};
        \addlegendentry{\hphantom{i}OSVOS-S \cite{maninis2017video}}
        
        \addplot[blue,mark=o, mark size=3.5,only marks, line width=0.75]coordinates{(4.66,74.1)};
        \addlegendentry{\hphantom{i}DyeNet \cite{li2018video}}
                
        \addplot[cyan,mark=o, mark size=3.5,only marks, line width=0.75] coordinates{(0.26,66.7)};
        \addlegendentry{\hphantom{i}RGMP \cite{oh2018fast}}

        \addplot[orange,mark=o, mark size=3.5,only marks, line width=0.75] coordinates{(0.28,54.8)};
        \addlegendentry{\hphantom{i}OSMN \cite{Yang2018osmn}}
                
        \addplot[magenta,mark=o, mark size=3.5,only marks, line width=0.75] coordinates{(3.6,58.2)};
        \draw (37.4,77.8) -- (5.02,73.7) -- (0.81,65.7);
        \addlegendentry{\hphantom{i}FAVOS \cite{Cheng_favos_2018}}
    \end{axis}
\end{tikzpicture}}
   \caption{Quality versus timing  on the DAVIS 2017 val set. For methods that only publish runtime results on the DAVIS 2016 dataset, we take these timings as per object timings and extrapolate to the number of objects in the DAVIS 2017 val set.}
   \label{fig:speed}
\end{figure}
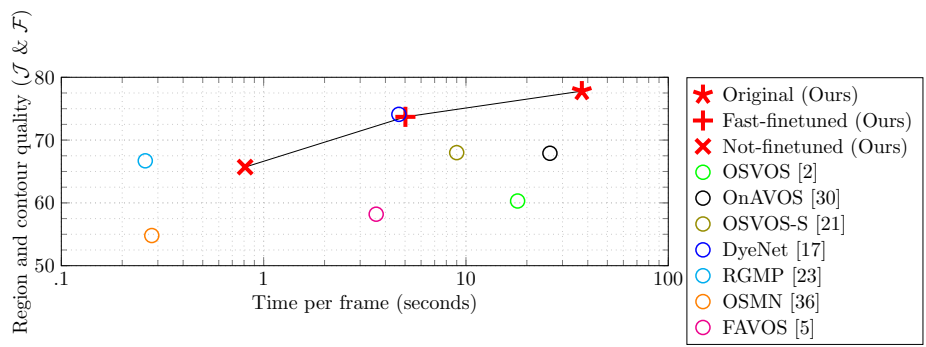

\subsection{Further Large-scale Evaluation}

In Table \ref{table:ytvos} we present the results of PReMVOS on the new YouTube-VOS dataset \cite{xu2018youtubemethod}, the largest VOS dataset to date (508 test video compared to 30 in DAVIS), our results on the test set obtained 1st place in the 1st Large-scale Video Object Segmentation Challenge. We don't run the original PReMVOS method but the PReMVOS Fast-finetuned version (see Section \ref{sec:runtime}), which can be evaluated on this larger dataset in a more reasonable amount of time. Our results are much better than \cite{xu2018youtubemethod}, the only other method that has published results on this dataset. It is also better than all other methods that submitted results to the 2017 1st Large-scale Video Object Segmentation Challenge.

\begin{table}[t!]
\newcolumntype{Y}{>{\centering\arraybackslash\hsize=0.16\hsize}X}
\newcolumntype{Z}{>{\centering\arraybackslash\hsize=0.11\hsize}X}
\newcolumntype{S}{>{\centering\arraybackslash\hsize=0.16\hsize}X}

\newcommand{\mmr}{\arrayrulecolor{lightgray}\cmidrule[0.25pt]{1-6}\arrayrulecolor{black}}
\setlength{\tabcolsep}{0pt}
\begin{center}
\begin{tabularx}{\textwidth}{SYYYYY}
\toprule[1pt]
 & Overall & $\mathcal{J}$ seen & $\mathcal{J}$ unseen & $\mathcal{F}$ seen & $\mathcal{F}$ unseen \\
\midrule[0.5pt]
Ours & \textbf{72.2} & \textbf{73.7} & 64.8 & \textbf{77.8} & 72.5\\
\mmr
Seq2Seq \cite{xu2018youtubemethod} & 70.0 & 66.9 & \textbf{66.8} & 74.1 & 72.3\\
\mmr
2nd & 72.0 & 72.5 & 66.3 & 75.2 & \textbf{74.1}\\
\mmr
3rd & 69.9 & 73.6 & 62.1 & 75.5 & 68.4\\
\mmr
4th & 68.4 & 70.6 & 62.3 & 72.8 & 67.7\\
\bottomrule[1pt]
\end{tabularx}
\end{center}
\caption{Results on the YouTube-VOS dataset, using the PReMVOS Fast-finetuned version of PReMVOS. These results obtained 1st place in the the 1st Large-scale Video Object Segmentation Challenge. '2nd','3rd','4th' refers to the other competitors results in this challenge with that ranking. Bold results are the best results for that metric.}
\label{table:ytvos}
\end{table}

%------------------------------------------------------------------------
\section{Conclusion}
In this paper we have presented a new approach for solving the video object segmentation task. Our proposed approach works by dividing this task into first generating a set of accurate object segmentation mask proposals for each video frame and then selecting and merging these proposals into accurate and temporally consistent pixel-wise object tracks over a video sequence. We have developed a novel approach for each of these sub-problems and have combined these into the PReMVOS (Proposal-generation, Refinement and Merging for Video Object Segmentation) algorithm. We show that this method is particularly well suited for the difficult multi-object video object segmentation task and that it produces results better than all current state-of-the-art results for semi-supervised video object segmentation on the DAVIS benchmarks, as well as getting the best score in the 2018 DAVIS Challenge.\\

\newcommand{\PAR}[1]{\noindent {\bf #1~}}

\noindent{\bf Acknowledgements.~} This project was funded, in parts, by ERC Consolidator Grant DeeViSe
(ERC-2017-COG-773161).

\newpage


 \bibliographystyle{splncs04}
 \bibliography{abbrev_short.bib,mybibliography}

\end{document}